\documentclass[conference]{IEEEtran}

\IEEEoverridecommandlockouts
\usepackage{cite}
\usepackage{pifont}
\usepackage{multirow}
\usepackage{amsmath,amssymb,amsfonts}
\usepackage{algorithmic}
\usepackage{graphicx}
\usepackage{textcomp}
\usepackage{xcolor}
\def\BibTeX{{\rm B\kern-.05em{\sc i\kern-.025em b}\kern-.08em
    T\kern-.1667em\lower.7ex\hbox{E}\kern-.125emX}}

\begin{document}

\title{\centering\makebox[\linewidth][c]{\parbox{1.1\textwidth}{\centering LAIP: Learning Local Alignment from Image-Phrase Modeling for Text-based Person Search}}\\
\thanks{This work is supported by Project Funded by the Priority Academic Program Development of Jiangsu Higher Education Institutions, and is also supported by the Open Projects Program of State Key Laboratory of Multimodal Artificial Intelligence Systems.}
}
\author{
\IEEEauthorblockN{Haiguang Wang$^{\dag}$\quad Yu Wu$^{\dag}$\quad Mengxia Wu\quad Min Cao$^{*}$\quad Min Zhang}
\IEEEauthorblockA{
\textit{School of Computer Science and Technology, Soochow University} \\
Suzhou, China \\
\{hgwang.cv, ywu.barry\}@gmail.com, mxwuwumx@stu.suda.edu.cn, 
\\
caomin0719@126.com,
minzhang@suda.edu.cn
\thanks{$^\dag$Equal contribution; $^*$Corresponding author}
}
}

\maketitle

\begin{abstract}
Text-based person search aims at retrieving images of a particular person based on a given textual description. A common solution for this task is to directly match the entire images and texts, \emph{i.e.,} global alignment, which fails to deal with discerning specific details that discriminate against appearance-similar people. As a result, some works shift their attention towards local alignment.
One group matches fine-grained parts using forward attention weights of the transformer yet underutilizes information. Another implicitly conducts local alignment by reconstructing masked parts based on unmasked context yet with a biased masking strategy. All limit performance improvement. This paper proposes the \textbf{L}ocal \textbf{A}lignment from \textbf{I}mage-\textbf{P}hrase modeling (LAIP) framework, with \textbf{Bidir}ectional \textbf{Att}ention-weighted local alignment (BidirAtt) and \textbf{M}ask \textbf{P}hrase \textbf{M}odeling (MPM) module.
BidirAtt goes beyond the typical forward attention by considering the gradient of the transformer as backward attention, utilizing two-sided information for local alignment. MPM focuses on mask reconstruction within the noun phrase rather than the entire text, ensuring an unbiased masking strategy. Extensive experiments conducted on the CUHK-PEDES, ICFG-PEDES, and RSTPReid datasets demonstrate the superiority of the LAIP framework over existing methods.
\end{abstract}

\begin{IEEEkeywords}
Text-based Person Search; Text-image Retrieval; Attention Mechanism; Gradient Information; Mask Strategy
\end{IEEEkeywords}

\section{Introduction}
Text-based person search (TBPS)~\cite{li2017person} seeks to retrieve images of a specific pedestrian with a given textual description of that person, which has drawn increasing attention in recent years. 
Researchers have proposed various methods
\footnote{A detailed introduction to the related work of TPBS can be found in the Appendix A.} 
to tackle this task over the years, resulting in impressive performance~\cite{li2017person, niu2020improving, wang2022caibc, bai2023text}. 
However, some critical issues remain underexplored yet, hindering accuracy improvements. 

\begin{figure}[t]
    \centering
    \setlength{\intextsep}{-1pt}
    \setlength{\abovecaptionskip}{-0.01cm}
    \includegraphics[width=0.95\linewidth]{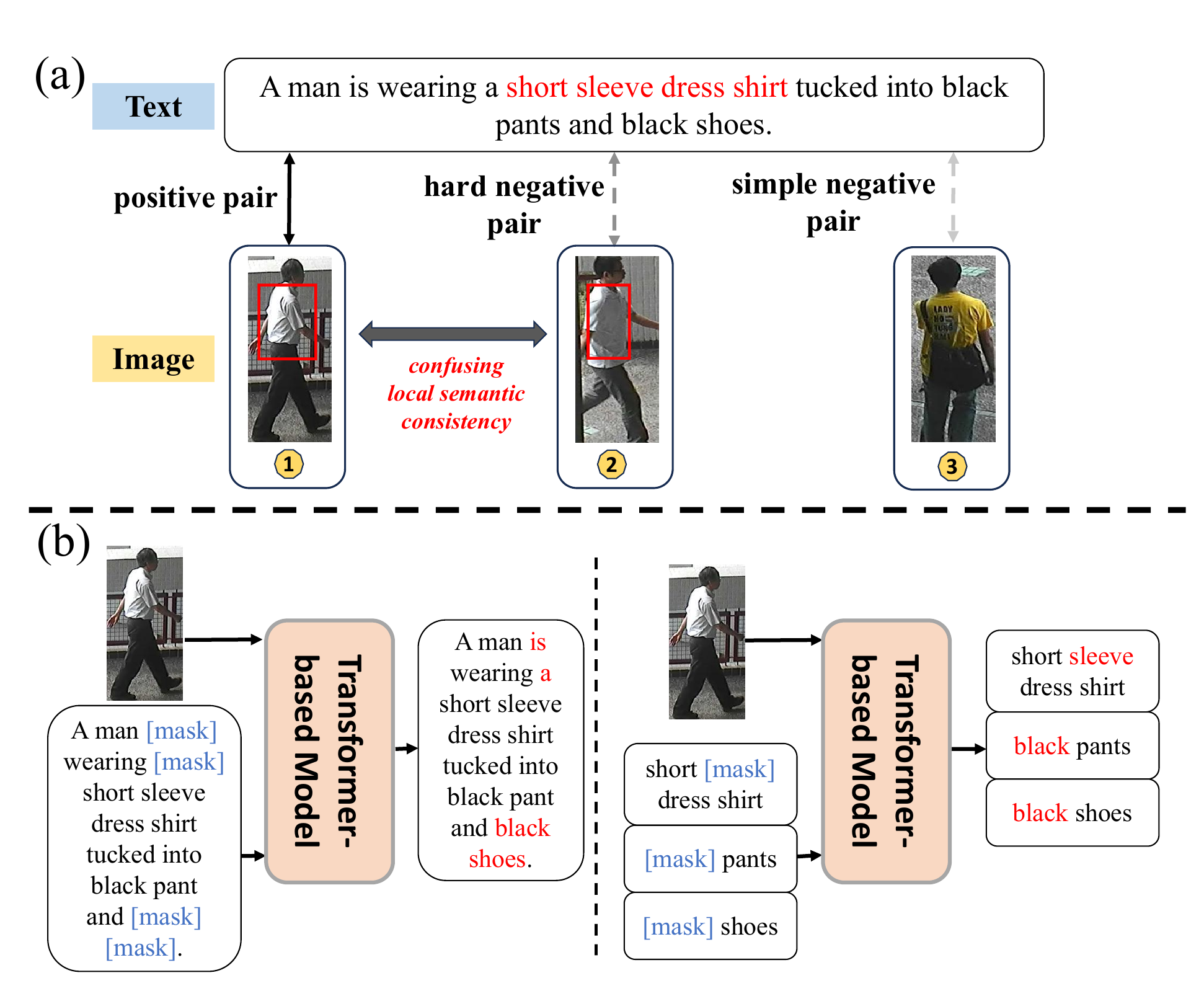}
    \caption{
    (a): Illustration of the \emph{confusing local semantic consistency} in text-based person search.
    (b): Comparison between MLM (left) and MPM (right).}
    \label{fig:simple_hard_sample}
\end{figure}

Especially, the challenges arise from the similar local physical appearances of individuals, leading to what is referred to as \emph{confusing local semantic consistency} among different persons.
As shown in Fig.~\ref{fig:simple_hard_sample} (a), grounded on a positive pair \textless text, image\#1\textgreater, the negative pair \textless text, image\#2\textgreater\space is highlighted in red box since two different persons share a common physiognomic feature (\emph{i.e.,} both wear 'a short sleeve dress shirt'). 
To distinguish the small difference caused by \emph{confusing local semantic consistency}, the model is required to be capable of bridging the non-linear gap to align the corresponding fine-grained information across modalities.

One feasible way is to seek improved cross-modal local alignment beyond the global alignment. 
These local alignment methods can be generally classified into two paradigms.
The first one~\cite{wang2020vitaa, niu2020improving, li2022learning} performs the local matching in an explicit way. 
The information from both modalities is divided into several parts. 
Subsequently, each part is rendered as the query, and the model proceeds to search for their corresponding counterparts from another modality in a hard~\cite{wang2020vitaa} or soft~\cite{niu2020improving, li2022learning} way. 
The second paradigm~\cite{wu2021lapscore, irra, Fujii_2023_ICCV} performs local alignment implicitly through reconstructing masked local features using context from unmasked areas. 

\begin{figure}
    \centering
    \setlength{\intextsep}{-1pt}
    \setlength{\abovecaptionskip}{-0.01cm}
    \setlength{\belowcaptionskip}{1cm}
    \includegraphics[width=0.95\linewidth, height=0.45\linewidth]{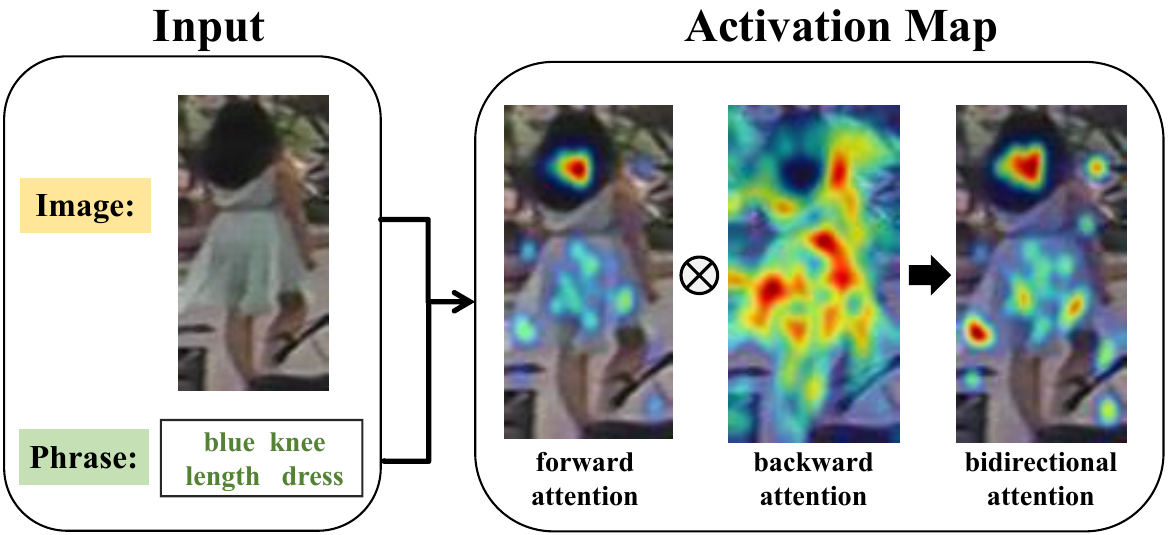}
    \caption{Visualization of the activation map of LAIP. The warmer the color tone, the stronger its activation to the input phrase.}
    \label{fig:gradcam_hotmap}
\end{figure}

\begin{figure*}[!htbp]
    \centering
    \setlength{\intextsep}{-1pt}
    \setlength{\abovecaptionskip}{0.01cm}
    \setlength{\belowcaptionskip}{-0.3cm}
    \includegraphics[width=\linewidth]{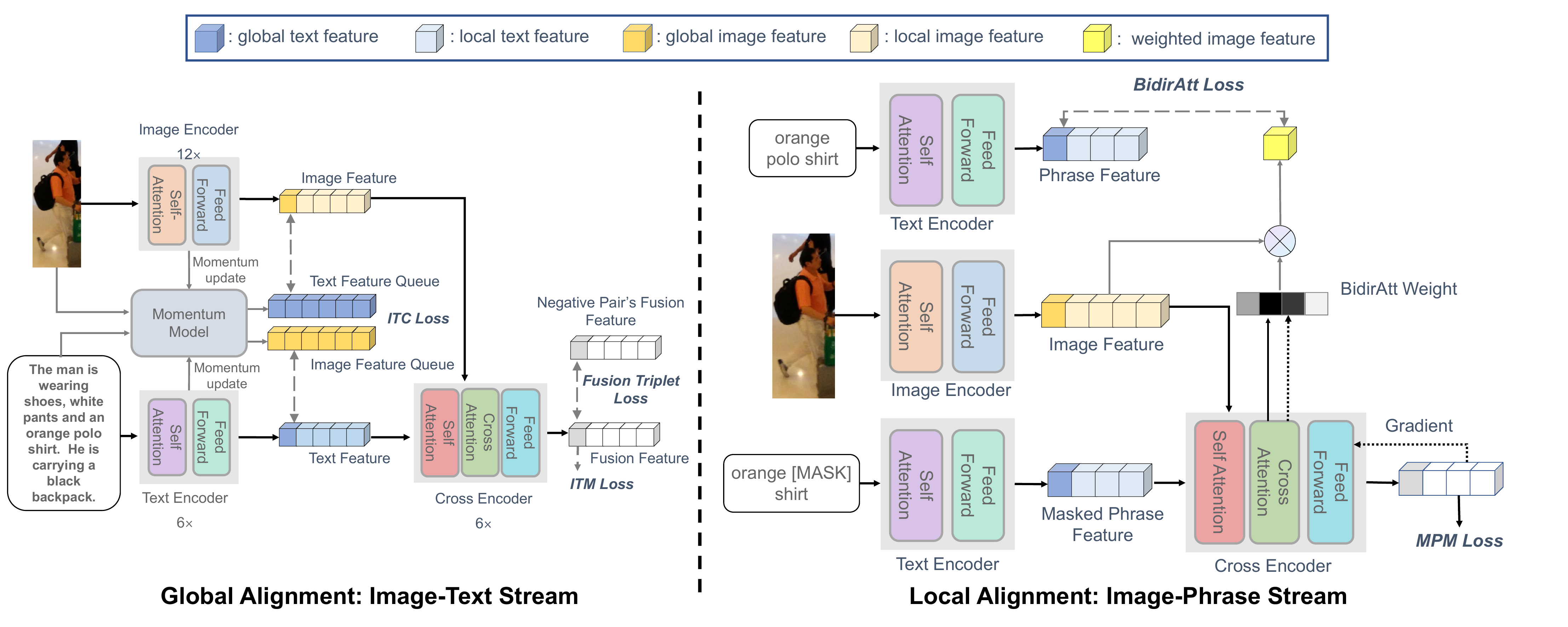}
    \caption{Illustration of the overall architecture. The left-hand side of this figure is the baseline ALBEF~\cite{li2021align} with additional fusion triplet loss
    for global alignment between the image and text. The right-hand side is the proposed BidirAtt+MPM for local matching between the image and phrase in the text.}
    \label{fig:model_architecture}
\end{figure*}
While local alignment methods can partially address the issue of \emph{confusing local semantic consistency}, certain challenges persist.
The first paradigm typically aligns local features across modalities by utilizing attention weights in the cross-modal transformer. 
These weights convey the relationship between modalities only in a forward process, providing one-sided information for local alignment. 
Beyond that, backward attention derived from the model's output gradient can be regarded as supplementary local information. As shown in Fig.~\ref{fig:gradcam_hotmap}, after considering backward attention, more emphasis compared with the activation map of forward attention is placed on the described object 'blue knee length dress'. 
For this, we consider incorporating forward and backward attentions for local alignment.
The second group encounters the dilemma that the masking strategy tends to be biased and thus affects performance. 
Since both modalities contain some inessential parts, such as conjunctions in text and backgrounds in the image, and reconstructing these parts brings no benefits to the capability of the model's local alignment. 
As shown in the left of Fig.~\ref{fig:simple_hard_sample} (b), reconstructing the words `is' and `a' serves no purpose in local alignment. 
For this, we employ the masked word reconstruction only at the phrase level instead of the sentence level which have rich semantic information.

In summary, we propose a Local Alignment from Image-Phrase modeling (LAIP) framework, including a \textbf{Bidir}ectional \textbf{Att}ention-weighted local alignment~(BidirAtt) and a \textbf{M}ask \textbf{P}hrase \textbf{M}odeling (MPM). We adopt the cross-modal pretrained model ALBEF~\cite{li2021align} as the baseline primarily oriented to global alignment. 
Beyond that, BidirAtt and MPM are added for local alignment. BidirAtt incorporates forward and backward attentions for local alignment. MPM, compared with Masked Language Model~(MLM)~\cite{Devlin_Chang_Lee_Toutanova_2019}, achieves local alignment implicitly by reconstructing the mask word within the noun phrase instead of the entire sentence.

Our contributions can be summarized as three-fold:
\begin{itemize}
    \setlength{\itemsep}{-2pt}
    \setlength{\parskip}{2pt}
    \item We consider the gradient information for the local alignment in TBPS for the first time, and propose a \textbf{Bidir}ectional \textbf{Att}ention-weighted local alignment (BidirAtt) module to enhance the explicit local alignment.
    \item We introduce a \textbf{M}ask \textbf{P}hrase \textbf{M}odeling (MPM) module, in which the crucial clues in texts tend to be masked, thus encouraging the model to focus on significant information for local alignment.
    \item Experiments on CUHK-PEDES, ICFG-PEDES and RSTPReid datasets prove that LAIP shows the competitive results compared to other methods.
\end{itemize}

\section{Method}
In this section, we elaborate on the proposed LAIP. The overall architecture is shown in Fig.~\ref{fig:model_architecture}.
This section first introduces the global alignment in the image-text stream, and then delineates the local alignment in the image-phrase stream, including the proposed BidirAtt and MPM.
Finally, we describe the training and inference.

\subsection{Global Alignment: Image-Text Stream}
We employ ALBEF~\cite{li2021align} as our baseline for global alignment.
It is a transformer-based model architecture with image and text encoders, the corresponding momentum model (a moving-average version of the unimodal encoders) and one cross-modal encoder. We use the image-text contrastive loss ($\mathcal{L}_{itc}$) and the image-text matching loss ($\mathcal{L}_{itm}$) for global alignment, which are used in ALBEF for Image-Text Retrieval. We further enhance global alignment by adding a fusion triplet loss ($\mathcal{L}_{tri}$) that enables contrast between positive and negative text-image pairs.More details can be found in appendix B and C.



Overall, in global alignment, we define the global loss as:
\begin{equation}
\mathcal{L}_{global}=\mathcal{L}_{itc}+\mathcal{L}_{itm}+\mathcal{L}_{tri}.\\ 
\end{equation}


\subsection{Local Alignment: Image-Phrase Stream}\label{image-phrase_stream}
To further excavate the fine-grained information, we center on local alignment to provide the potential for aligning image patches and textual phrases.
For this, we add an image-phrase stream into the overall architecture, mainly including the proposed BidirAtt and MPM. 


Following existing works~\cite{niu2020improving, shu2022see}, we leverage the off-the-shelf Natural Language ToolKit~(NLTK)~\cite{loper2002nltk} to parse noun phrases in the text. The noun phrases are defined according to the part-of-speeches and context of the word, \emph{e.g.}, black hair~(adjective+noun). Given the input image-phrase pair $(I, P)$, we obtain image representations $F^I=\{f^I, f^I_{1}, f^I_{2}, ..., f^I_{L_I}\} \in \mathbb{R}^{(L_I+1)\times d}$ by the image encoder and phrase representations $F^{P}=\{f^{P}, f^{P}_1, \cdots , f^{P}_{L_{P}}\}\in \mathbb{R}^{(L_P+1)\times d}$ by text encoder.
The image-phrase fusion representations $F^U=\{f^ U, f^U_{1}, f^U_{2}, \cdots , f^U_{L_P}\}\in \mathbb{R}^{(L_P+1)\times d}$ can be computed by inputting the image and phrase representations into the cross-modal encoder, which is as same as one in the image-text stream. $f^I$, $f^P$ and $f^U$ denote the global representations of the image, phrase and image-phrase fusion, respectively.
$f^I_{i}$ $(i=1, 2, \cdots, L_I)$ and $f^P_{j}$ $(j=1, 2, \cdots, L_P)$ indicate the representations of the image patches and textual pharse tokens, respectively. 

\subsubsection{Bidirectional Attention-weighted Local Alignment}
The forward attention in the cross-modal transformer directly captures the relationship between tokens in different modalities through attention weights.
Beyond that, the gradient of the prediction probability from the cross-modal transformer also provides valuable information.
There have been several methods of visual explanation~\cite{gradcam, gradcam++} using gradient information.
Inspired by it, we use the gradient as a backward attention, together with the forward attention into the bidirectional Attention for local alignment.

\begin{figure}
    \centering
    \setlength{\intextsep}{-1pt}
  \setlength{\abovecaptionskip}{0.01cm}
  \setlength{\belowcaptionskip}{-0.65cm}
    \includegraphics[width=0.9\linewidth]{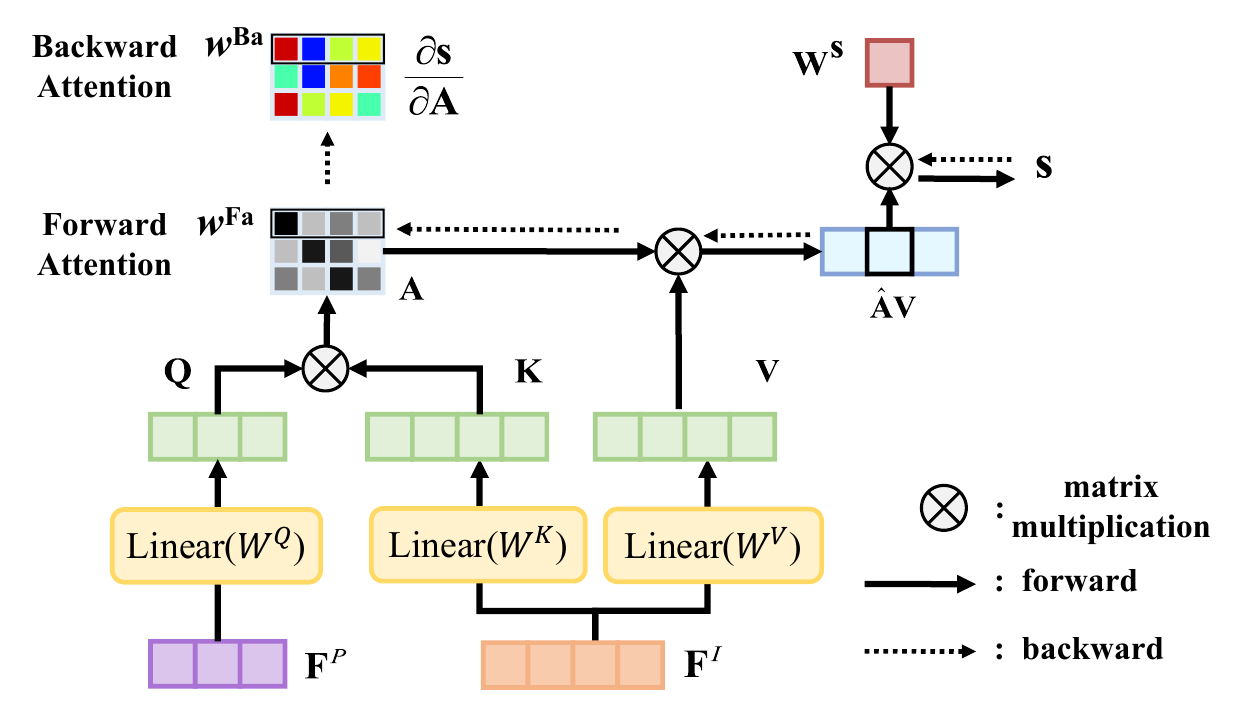}
    \caption{A simplified computation procedure of the forward attention and backward attention.}
    \label{fig:channel_wise_attention}
\end{figure}

To make the analysis of bidirectional Attention more clear, we deduce the formulations of the forward attention and backward attention only in a cross attention layer, and also illustrate their computation procedures in Fig.~\ref{fig:channel_wise_attention}. 

Starting from the phrase representations $F^P$ and image representations $F^I$ as the input of the cross-modal encoder, we present its internal computation process by taking one head of a cross attention layer as an example,
\begin{equation}
   \begin{aligned}
        Q=F^PW^Q,
        K=F^IW^K,
        V=F^IW^V,
    \end{aligned} 
\end{equation}
\begin{equation}
    A=\mathrm{softmax}(\frac{QK^T}{\sqrt{d'}}),
\label{eq:A}
\end{equation}
where $d'=\frac{d}{H}$ and $H$ is the head's number. $W^Q, W^K, W^V\in \mathbb{R}^{d \times d'}$ are linear projection matrices and $K^T$ is the transpose of $K$. 
$A\in \mathbb{R}^{(L_{P}+1)\times (L_{I}+1)}$ depicts the similarity relation between the phrase tokens and image patches. 
The \textbf{forward attention} of the representation of the $j$-th image patch  to the global representation of the phrase is
\begin{equation}
    w^{Fa}_j = A_{0j},
\end{equation}
where $A_{0j}$ is the first row, $j$-th column of A.

The prediction score of the masked word's label is
\begin{equation}
    s=(\hat{A}V)W^s,
    \label{eq:s}
\end{equation}
where $W^s\in \mathbb{R}^{d'\times 1}$ is a linear projection, $\hat{A}\in \mathbb{R}^{1\times (L_{I}+1)}$ denotes the representation of each image patch to the representation of the masked token in phrase from $A$.

The \textbf{backward attention} based on $s$ can be represented as:
\begin{equation}
  \begin{aligned}
  w^{Ba}_j  =\frac{\partial s}{\partial w_j^{Fa}}
        = \sum_{k=1}^{d'}V_{jk}\cdot W^s_k,
    \end{aligned}
    \label{eq:ba}
\end{equation}
which sheds light on the effect of $j$-th projected image patch's presentation $V_{j}$ on the final prediction result. The higher the $w^{Ba}_j$, the more the $j$-th image patch contributes to the decision-making process.
The backward attention  indirectly indicates the semantic relationship between $j$-th image patch and the phrase, as shown in Fig.~\ref{fig:gradcam_hotmap}. 

Finally, the \textbf{bidirectional attention} $w_j$ is formulated as:
\begin{equation}
  \begin{aligned}
  w_j=\text{norm}(\overline{w}^{Fa}_j \cdot \overline{w}^{Ba}_j),
    \end{aligned}
\end{equation}
where $\overline{w}^{Fa}_j$ and $\overline{w}^{Ba}_j$ are the average of $w^{Fa}_j$ and $w^{Ba}_j$ over all heads, respectively. $\text{norm}(\cdot)$ is a function that normalizes the weights and ensures their summation equals 1.


According to the bidirectional attention, 
we compute the weighted global representation by 
\begin{equation}
\widetilde{f^{I}}=\sum_{j=1}^{L_I}w_j\cdot f^I_j,
\end{equation}
where $w_j$ reflects the weight of $j$-th image patches to the input phrase.
As a result, $\widetilde{f^{I}}$, the sum of all weighted patches' representations, can also be deemed as the phrase-guided image global representation. The representations of image patches corresponding to the phrase (i.e., larger $w$) are highlighted, and others are weakened.

Now, we can compute the coarse similarity between image ${I}$ and phrase $P$ as:
\begin{equation}
    sim_{coarse}({I}, P)=\frac{\mathcal{G}_I(\widetilde{f^I})\cdot \mathcal{G}_P(f^P)}{\Vert \mathcal{G}_I(\widetilde{f^I})\Vert_2 \cdot \Vert \mathcal{G}_P(f^P)\Vert_2},
\end{equation}
where $\Vert\cdot\Vert_2$ denotes the norm of the representation, $\mathcal{G}_P$ (\emph{resp.} $ \mathcal{G}_I$) is the linear projection that maps $f^P$ (\emph{resp.} $\widetilde{f^I}$) to the representation with a lower dimension.

Overall, We carry out the bidirectional attention-weighted local alignment loss for one image-phrase pair as: 
\begin{equation}
\mathcal{L}_{biatt} =1- sim_{coarse}({I}, P).
\label{eq:biatt}
\end{equation}

\subsubsection{Mask Phrase Modeling}
Beyond the BidirAtt for explicit local alignment,
we propose a mask phrase modeling (MPM) for implicit local alignment.

Specifically, we execute the mask operation on the inputted phrase $P$ by randomly replacing one word in the phrase with the token [MASK].
The masked phrase is denoted as $\hat{P}$.
Inputting the pair $(I, \hat{P})$ into the model, we compute the final fusion representation $F^{\hat{U}}=\{f^{\hat{U}}, f^{\hat{U}}_{1}, f^{\hat{U}}_{2}, \cdots ,  f^{\hat{U}}_{L_P}\} \in \mathbb{R}^{(L_p+1)\times d}$.
Then, we leverage a multi-layer perception (MLP) classifier to predict the origin tokens in the phrase by the MPM loss of one image-phrase pair:
\begin{equation}
    \mathcal{L}_{mpm} = \sum_{j=1}^{Lp}\operatorname{CrossEntropy}(\mathrm{MLP}(f^{\hat{U}}_{j}), y_j), 
    \label{eq:mpm}
\end{equation}
where $y_j\in \mathbb{R}^{|\mathcal{V}|}$ is an one-hot vocabulary distribution of $j$-th token where the ground-truth token has probability of $1$ and others $0$. $|\mathcal{V}|$ is the size of the vocabulary. 
\subsection{Training and Inference}\label{prediction}
In the training phase, we optimize the overall losses as:
\begin{equation}
\mathcal{L}=\mathcal{L}_{global}
            +\sum_{ph\in {\mathcal{P}\mathcal{H}}}(\mathcal{L}_{biatt}^{ph}+\mathcal{L}_{mpm}^{ph}), 
\end{equation}
where $\mathcal{P}\mathcal{H}$ is the set of phrases in the input text.

\begin{table}[!h]
    \centering
    \caption{Performance comparison with the state-of-the-art methods on CUHK-PEDES dataset.}
    \resizebox{0.48\textwidth}{!}{
        \begin{tabular}{l|l|cccc}
            \hline
            Method & Reference & R@1 & R@5 & R@10 & mAP\\
            \hline
            GNA-RNN~\cite{li2017person} & CVPR17 & 19.05 & - & 53.64 & - \\
            DSSL~\cite{zhu2021dssl} &  MM21 & 59.98 & 80.41 & 87.56 & - \\
            LapsCore~\cite{wu2021lapscore} & ICCV21 & 63.40 & - & 87.80& - \\
            SAF~\cite{li2022learning} & ICASSP22 & 64.13 & 82.62 & 88.40 & 58.61 \\
            CAIBC~\cite{wang2022caibc} & MM22 & 64.43 & 82.87 & 88.37 & - \\
            IVT~\cite{shu2022see} & ECCVW22 & 65.59 & 83.11 & 89.21 & 60.66 \\
            $C_2A_2$~\cite{c2a2} &  MM22 & 67.94 & 86.86 & 91.87 & - \\
            UniPT~\cite{Shao_2023_ICCV} & ICCV23 & 68.50 & 84.67 & - & - \\
            IRRA~\cite{irra} & CVPR23 & 73.38 & 89.93 & \textbf{93.71} & \textbf{66.13} \\
            \hline
            ALBEF~\cite{li2021align} & NeurIPS21 & 63.87 & 81.73 & 86.42 & 54.41 \\
            LAIP (Ours) & - & \textbf{76.72} & \textbf{90.42} & 93.60 & 66.05 \\
            \hline
        \end{tabular}}
    \label{tab:compare_with_sota_cuhkpedes}
\end{table}
\begin{table}[htbp]
    \centering
    \caption{Performance comparison with the state-of-the-art methods on ICFG-PEDES dataset.}
    \resizebox{0.48\textwidth}{!}{
        \begin{tabular}{l|l|cccc}
            \hline
            Method & Reference & R@1 & R@5 & R@10 & mAP\\
            \hline
            Dual-path~\cite{zheng2020dual} & TOMM20 & 38.99 & 59.44 & 68.41 & - \\
            CMPM/C~\cite{zhang2018deep} & ECCV18 & 43.51 & 65.44 & 74.26 & - \\
            TIPCB~\cite{chen2022tipcb} & Neuro22 & 54.96 & 74.72 & 81.89 & - \\
            IVT~\cite{shu2022see} & ECCVW22 & 56.04 & 73.60 & 80.22 & -  \\
            UniPT~\cite{Shao_2023_ICCV} & ICCV23 & 60.09 & 76.19 & - & - \\
            CFine~\cite{yan2022clip} &arXiv22  & 60.83 & 76.55 & 82.42 & - \\
            IRRA~\cite{irra} & CVPR23 & 63.46 & \textbf{80.25} & \textbf{85.82} & \textbf{38.06} \\
            \hline
            ALBEF~\cite{li2021align} & NeurIPS21 & 44.36  & 60.79  & 67.28 & 19.98 \\
            LAIP (Ours) & - & \textbf{63.52} & 79.28  & 84.57 & 37.02  \\
            \hline
        \end{tabular}}
    \label{tab:compare_with_sota_icfg_pedes}
\end{table}

\begin{table}[!htbp]
    \centering
    \caption{Performance comparison with the state-of-the-art methods on RSTPReid dataset.}
    \resizebox{0.48\textwidth}{!}{
        \begin{tabular}{l|l|cccc}
            \hline
            Method & Reference & R@1 & R@5 & R@10 & mAP \\
            \hline
            DSSL~\cite{zhu2021dssl} & MM21 & 39.05 & 62.6 & 73.95 &- \\
            IVT~\cite{shu2022see} & ECCVW22 & 46.70 & 70.00 & 78.80 & -\\
            CFine~\cite{yan2022clip} & arXiv22 & 50.55 & 72.50 & 81.60 &-\\
            IRRA~\cite{irra} & CVPR23 & 60.20 & 81.30 & 88.20 & \textbf{47.17} \\
            \hline
            ALBEF~\cite{li2021align} & NeurIPS21 & 58.60 & 79.60 & 85.15 & 40.50 \\
            LAIP (Ours) & - & \textbf{62.00} & \textbf{83.15}  & \textbf{88.50} & 45.27 \\
            \hline
        \end{tabular}}
    \label{tab:compare_with_sota_rstp_reid}
\end{table}
In the inference phase, we employ the candidate selection strategy for accelerating inference with further details provided in the Appendix D. It is notable that the BidirAtt and MPM are only employed during the training phase and are discarded in the inference stage.
\begin{table*}[!htbp]
    \centering
    \setlength{\abovecaptionskip}{0.15cm}
    \setlength{\belowcaptionskip}{-0.4cm}
    \caption{Ablation study on each component on CUHK-PEDES. "ITC" and "ITM" denotes $\mathcal{L}_{itc}$ and $\mathcal{L}_{itm}$, "ForAtt" and "BackAtt" are abbreviations for Forward Attention and Backward Attention, respectively. "$Ftri$" is fusion triplet loss $\mathcal{L}_{tri}$. }
    \resizebox{0.98\textwidth}{!}{
        \begin{tabular}{c|l|c|cc|ccc|cccc}
            \hline
            \multirow{3}{*}{\textbf{No.}}&\multirow{3}{*}{\textbf{Methods}}  & \multirow{3}{*}{\textbf{COCO-finetune}} & 
            \multicolumn{2}{c|}{\textbf{Global Matching}} & \multicolumn{3}{c|}{\textbf{Local Matching}} & \multirow{3}{*}{\textbf{R@1}} & \multirow{3}{*}{\textbf{R@5}} & \multirow{3}{*}{\textbf{R@10}} & \multirow{3}{*}{\textbf{mAP}} \\
            \cline{4-8}
            & & & \multirow{2}{*}{\textbf{ITC+ITM}} & \multirow{2}{*}{\textbf{Ftri}} & \multicolumn{2}{|c}{\textbf{BidirAtt}} &  \multirow{2}{*}{\textbf{MPM}} & &  &  &  \\
            & & & & &\textbf{ForAtt} & \textbf{BackAtt} & &  &  &  &  \\
            \hline
            1&Baseline & & $\checkmark$ & & & & & 63.87 & 81.73 & 86.42 & 54.41 \\
            2&Baseline$^+$ & $\checkmark$ & $\checkmark$ & & & & & 66.59 & 84.31 & 88.99 & 57.08 \\
            \hline
            3&+ForAtt & $\checkmark$ & $\checkmark$ &  & $\checkmark$ &  & & 75.05 & 89.05 & 92.61 & 64.22  \\
            4&+BackAtt & $\checkmark$ & $\checkmark$ &  & & $\checkmark$ &  & 75.37 & 89.54 & 92.82 & 65.10  \\
            \hline
            5&+BidirAtts & $\checkmark$ & $\checkmark$ &  & $\checkmark$ & $\checkmark$ &  & 75.52 & 89.80 & 93.70  & 66.07 \\
            6&+MPM & $\checkmark$ & $\checkmark$ &  &  & & $\checkmark$ & 75.45 & 90.37 & \textbf{93.73} & \textbf{66.39}  \\
            7&+BidirAtts+MPM & $\checkmark$ & $\checkmark$ &  & $\checkmark$ & $\checkmark$ & $\checkmark$ & 75.15 & 90.04 & 93.63 & 66.02  \\
            8&LAIP\textbf{(Ours)} & $\checkmark$ & $\checkmark$ & $\checkmark$ & $\checkmark$ &$\checkmark$ & $\checkmark$ & \textbf{76.72} & \textbf{90.42} & 93.60 & 66.05  \\
            \hline
        \end{tabular}
    }
    \label{tab:ablation_study_component}
\end{table*}
\begin{table}[!htbp]
    \centering
    \caption{Comparison MPM with MLM on LAIP on CUHK-PEDES.}
    \resizebox{0.48\textwidth}{!}{
        \begin{tabular}{l|cccc}
            \hline
            Method & R@1 & R@5 & R@10 & mAP \\
            \hline
            LAIP (w/o MPM) & 75.21 & 89.53  & 93.19 & 65.91 \\
            LAIP (MLM) & 76.38 & 90.27  & \textbf{93.79} & 65.94 \\
            LAIP (MPM) & \textbf{76.72} & \textbf{90.42} & 93.60 & \textbf{66.05} \\
            \hline
        \end{tabular}}
    \label{tab:ablation_study_mpm}
\end{table}
\section{Experiments}
This section presents the experimental results of the proposed LAIP on three text-based person search datasets, i.e., CUHK-PEDES~\cite{Li:2008:PUC:1358628.1358946}, ICFG-PEDES~\cite{ding2021semantically}, and RSTPReid~\cite{zhu2021dssl}.
In addition, We also demonstrate the effectiveness of the LAIP via several ablation studies and visualization.

\subsection{Implementation Details}
All experiments are conducted on 4 NVIDIA 3090 GPUs. 
We initialize the parameters of LAIP using the ALBEF~\cite{li2021align}, whose weights are initially pre-trained on large-scale pre-training datasets and further finetuned on the COCO~\cite{coco} dataset. 
Subsequently, the LAIP is trained in a two-stage way on the TBPS datasets.
During the first stage, it is trained for 30 epochs with only $\mathcal{L}_{itc}$ and $\mathcal{L}_{itm}$ losses. 
In the second stage, it is trained for 15 epochs with all losses. 
More implementation details are covered in Appendix E.

\subsection{Comparison with State-of-the-art Methods}
Table \ref{tab:compare_with_sota_cuhkpedes}, Table \ref{tab:compare_with_sota_icfg_pedes} and Table \ref{tab:compare_with_sota_rstp_reid} provide a comprehensive performance comparison between the proposed LAIP method and state-of-the-art approaches on the CUHK-PEDES, ICFG-PEDES, and RSTPReid datasets, respectively. Notably, LAIP outperforms the baseline ALBEF across all evaluation metrics, demonstrating significant improvements. Furthermore, compared to recent state-of-the-art method IRRA~\cite{irra}, LAIP achieves decent advancement, showcasing its competitiveness.

\subsection{Ablation Study}\label{sec:ablation}
To evaluate the effectiveness of each component in LAIP, we conduct thorough ablation experiments on CUHK-PEDES, as shown in Table~\ref{tab:ablation_study_component}.
More analysis of hyper-parameters and visualization are in Appendix E, respectively.


\textbf{Effectiveness of BidirAtt.}
The BidirAtt integrates forward and backward attention weights for local alignment.
As presented in No.2, No.3, and No.4 of Table~\ref{tab:ablation_study_component}, compared with the baseline$^+$,
adding individual ForAtt and BackAtt both achieve the remarkable performance of 8.46\% and 8.78\% in terms of R@1, respectively, which highlight the importance of local alignment for text-based person search.
We note that adding the BackAtt yields modestly better performance than the ForAtt, with gains of 0.32\% and 0.88\% on R@1 and mAP, respectively.
The results indicate that backward attention also plays an important role in inter-modal alignment.
When combining forward and backward attention in BidirAtt, we harvest further enhanced performance of 75.52\% on R@1 and 66.07\% on mAP, demonstrating the effectiveness of BidirAtt.

\textbf{Effectiveness of MPM.}
The MPM module is designed for implicit local alignment by reconstructing the masked words within the noun phrases.
As shown in No.2 and No.6 of Table~\ref{tab:ablation_study_component}, with the effectiveness of the MPM module, the model achieves a significant performance improvement of 8.86\% and 9.31\% on R@1 and mAP, respectively.
Moreover, we also conduct experiments to compare our MPM module with the existing Masked Language Model (MLM) module, which aims at reconstructing the masked words within the entire person's description.
From Table~\ref{tab:ablation_study_mpm}, LAIP with MPM gains a modest performance improvement of 0.34\% at R@1.
We attribute the improvement to the unbiased masking strategy of MPM, which helps capture key clues for local alignment.


\textbf{Effectiveness of combining BidirAtt with MPM.} 
While BidirAtt and MPM individually achieve impressive performance gains, directly combining them poses optimization conflicts, as evident in No.5, No.6, and No.7 of Table \ref{tab:ablation_study_component}. 
The brute force combination of BidirAtt and MPM results in drops of $0.37\%$ and $0.30\%$ in R@1 compared to BidirAtt only and MPM only, respectively. 
However, we further enhance global alignment by devising a fusion triplet loss that enables contrast between positive and negative text-image pairs. As shown in No.8, the addition of fusion triplet loss for global alignment yields a significant performance gain of 1.57\% in R@1 compared to the combination of BidirAtt and MPM in No.7. This improvement could be attributed to achieving a better trade-off between global and local alignment.

\subsection{Visualization}\label{app:visiualization}
\begin{figure*}[htbp]
    \centering
    \setlength{\intextsep}{-1pt}
    \setlength{\abovecaptionskip}{-0.01cm}
    \setlength{\belowcaptionskip}{1cm}
    \includegraphics[width=\linewidth]{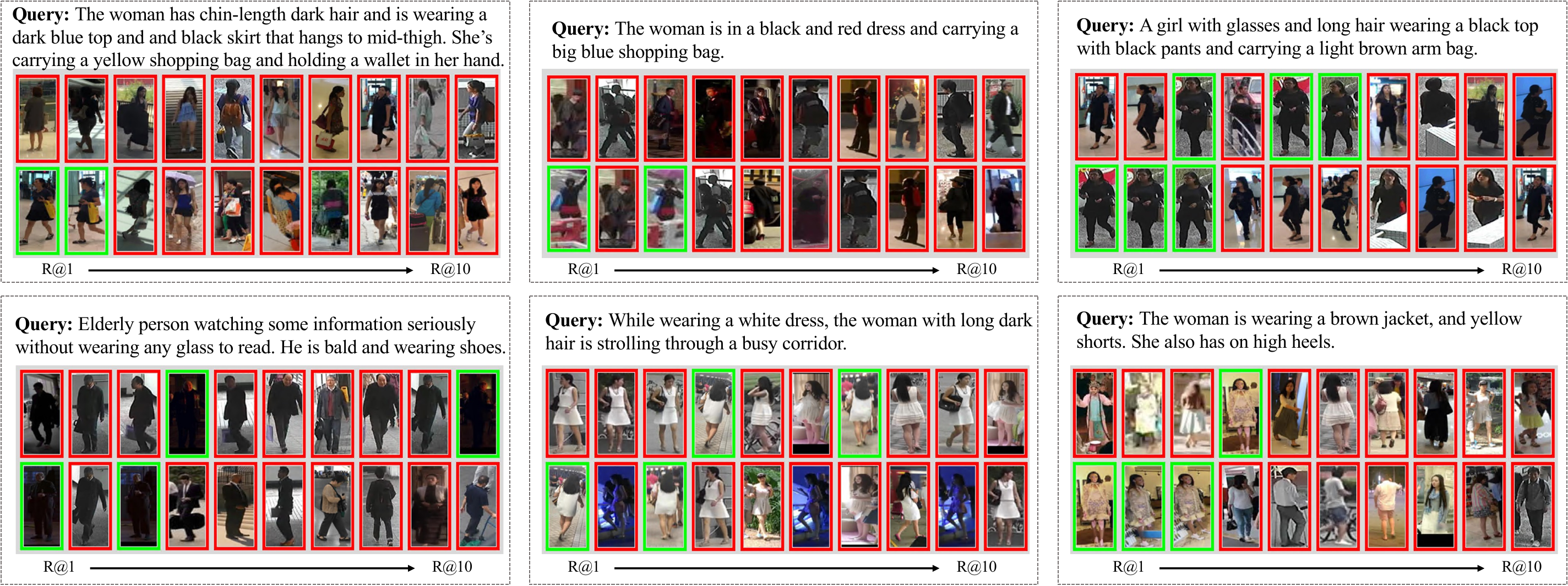}
    \caption{Visualization of top-10 retrieval results on CUHK-PEDES.
The first row in each example presents the retrieval results from the baseline ALBEF~\cite{li2021align}, and the second row shows the results from LAIP.
Correct/Incorrect images are marked by \textcolor{green}{green} / \textcolor{red}{red} rectangles.}
\label{fig:visualization}
\end{figure*}

\textbf{Visualization of the top-10 retrieval results on CUHK-PEDES.}
Fig.~\ref{fig:visualization} showcases diverse examples to illustrate the top-10 retrieval results of both the baseline ALBEF~\cite{li2021align} and the proposed LAIP. 
LAIP exhibits superior performance in retrieving positive samples compared to the baseline method. 
In these examples, the baseline becomes perplexed by the hard images that possess nearly identical appearances with the ground truths but vary in minor details. 
In the first example, the differences between the top $1$ retrieval result of the baseline and the ground truths are the `yellow shopping bag' and 'a wallet in hand', which are overlooked by the baseline and thus leading to false results, while LAIP captures such fine-grained details and succeeds in searching the correct person. 
This advancement is ascribed to the proposed BidirAtt and MPM, which enable capturing more precise alignment at the fine-grained level. The visualization serve as compelling evidence, affirming the prowess of LAIP in local alignment.
\section{Conclusion}
In this paper, we propose a learning Local Alignment from Image-Phrase modeling (LAIP) framework, which facilitates joint learning of global and local alignments, and boost the performance by optimising image-phrase aligning. To this end, we present two new modules: 1) Bidirectional Attention-weighted local alignment (BidirAtt) which integrates forward and backward attention weights to filter out semantically unrelated image patches to the textual phrase for local alignment; 2) Mask Phrase Modeling (MPM) that reconstructs highly informative word in phrase to further foster local alignment. 
LAIP achieves notable performance on CUHK-PEDES, ICFG-PEDES and RSTPReid benchmarks. 

\appendix
\label{app:introduction}
\subsection{Text-based Person Search}
\label{app:related_work_tbps}
The text-based person search task, initially proposed by Li \emph{et al.}~\cite{li2017person}, is notably more intricate due to its fine-grained nature compared with general cross-modal retrieval tasks. 
Early works~\cite{zhang2018deep, zheng2020dual, han2021text} tackle this task through global alignment methods, whereby the text and image are separately encoded, followed by the computation of similarity between these encoded features.
To extract more discriminative features for global alignment, Zheng \emph{et al.}~\cite{zheng2020dual} propose an instance loss ; Zhang \emph{et al.}~\cite{zhang2018deep} propose a cross-modal projection matching (CMPM) loss and a cross-modal projection classification (CMPC) loss.
Considering the fine-grained nature of the task, recent works mainly elaborate on designing local alignment methods, which could be divided into two diagrams, \emph{i.e.,} explicit and implicit local alignment.

In the explicit local alignment approaches~\cite{wang2020vitaa, li2022learning, zheng2020hierarchical, ding2021semantically, yan2022clip, 10219708, 10345496, 10253470}, the information from both images and texts is divided into several local parts, and then, the model proceeds to align the corresponding local parts from the two modalities.
Some works~\cite{wang2020vitaa, zheng2020hierarchical, ding2021semantically, yan2022clip, 10219708} directly split the text into words and images into patches or objects, then align these local components between texts and images.
In contrast, some works~\cite{li2022learning, 10345496, 10253470} conduct the local alignment in a more soft way that enables the adaptive selection and aggregation of features with similar semantics into local-aware features. 
However, all these methods only pay attention to the importance of forward propagation, neglecting the valuable information provided by gradients during backward propagation.
For the implicit ones, existing works~\cite{wu2021lapscore, irra, Fujii_2023_ICCV, 10.1145/3581783.3612244} carry out the local alignment by auxiliary reasoning tasks based on the local information.
Wu \emph{et al.}~\cite{wu2021lapscore} put forward a representation learning method based on color reasoning with the observation that color plays an important role in text-based person search.
Some works~\cite{irra, 10.1145/3581783.3612244} randomly mask words from entire texts, then reconstruct the masked feature by the unmasked contextual information.
Nevertheless, the masks of the unimportant words (e.g., 'is') have limited contribution to the inter-modal local alignment. 
In this paper, we propose Local Alignment from Image-Phrase modeling (LAIP) framework, which incorporates the gradient information with the weights of forward prorogation, and design a proxy task for unbiased mask reconstruction. 

\label{app:method}
Here, we describe the details of ALBEF, including its architecture, the global alignment stream, and candidate selection strategy during inference.

\subsection{ALBEF Architecture}
The unimodal encoders learn the feature representations of text and image, respectively, and the cross-modal encoder further learns the text-image pair’s representation by enabling interaction between the information of two modalities.
To be specific, ALBEF~\cite{li2021align} adopts ViT as the image encoder, and the first $6$ layers of BERT as the text encoder. It modifies the last $6$ layers of BERT by adding a cross-attention module and a linear projection module after the self-attention module in each layer as the cross-modal encoder.

\subsection{Global Alignment Stream}
We employ ALBEF~\cite{li2021align} as the baseline for global alignment. Given an input image-text pair $(I, T)$, we first feed them to the unimodal encoders, and obtain visual representations $F^I=\{f^I, f^I_{1}, f^I_{2}, ..., f^I_{L_I}\} \in \mathbb{R}^{(L_I+1)\times d}$ and textual representations $F^T=\{f^T, f^T_{1}, f^T_{2}, ..., f^T_{L_T}\} \in \mathbb{R}^{(L_T+1)\times d}$. $f^I$ and $f^T$ denote the global representations of the image and text, and $f^I_{i}$ $(i=1, 2, \cdots, L_I)$ and $f^T_{j}$ $(j=1, 2, \cdots, L_T)$ are the representations of image patches and textual tokens, respectively. $d$ is the dimension of representation.

Afterwards, we calculate the global embeddings of the text and the image in the following way:
\begin{align}
    v^{embed} = g_v(v_g),\\
    t^{embed} = g_t(t_g),
\end{align}
where $v^{embed}, t^{embed}\in \mathbb{R}^{d'}(d'<d)$ is for the calculation of \textbf{coarse similarities} between texts and images. To be concrete, we obtain the coarse similarities of text embed $t^{embed}$ and image embed $v^{embed}$ by calculating their cosine similarity:

Based on it, we can compute the coarse similarity between the image and text as:
\begin{equation}
    sim_{coarse}(I, T)=\frac{\mathcal{G}_I(f^I)\cdot \mathcal{G}_T(f^T)}{\Vert \mathcal{G}_I(f^I)\Vert_2 \cdot \Vert \mathcal{G}_T(f^T)\Vert_2},
\label{eq:coarse}
\end{equation}
where $\Vert\cdot\Vert_2$ denotes the norm of the representation, $\mathcal{G}_T$ (\emph{resp.} $ \mathcal{G}_I$) is the linear projection that maps $f^T$ (\emph{resp.} $f^I$) to the representation with a lower dimension.


Following the exponential moving average in MoCo~\cite{moco}, ALBEF introduces the momentum image encoder and momentum text one with the updated parameters $\theta^m \leftarrow \alpha\theta^m + (1-\alpha)\theta$. $\theta$ is the parameter of image/text encoder and $\alpha\in [0, 1)$ is a momentum coefficient. 
During training, recent $M_Q$ global representations of the image and text from the momentum encoders are stored in two queues $Q^{I}$ and $Q^{T}$, respectively.

By combining the representations both from the original encoder and the paired momentum one, we compute the image-to-text and text-to-image similarities as:  
\begin{equation}
    \begin{aligned}
        p^{i2t}(I, T^Q_j)=\frac{\mathrm{exp}(\mathrm{sim}_{coarse}(I, T^Q_j)/\tau)}{\sum_{i=1}^{M_Q} \mathrm{exp}(\mathrm{sim}_{coarse}(I, T^Q_i)/\tau)}, \\
        p^{t2i}(I^Q_j, T)=\frac{\mathrm{exp}(\mathrm{sim}_{coarse}(I^Q_j, T)/\tau)}{\sum_{i=1}^{M_Q} \mathrm{exp}(\mathrm{sim}_{coarse}(I^Q_i, T)/\tau)},
    \end{aligned}
\end{equation}
where $sim_{coarse}(I, T^Q_j)$ measures the coarse similarity between the image's global representation and the $j$-th text's global representation in $Q^T$, analogous to $sim_{coarse}(I^Q_j, T)$.
$\tau$ is the learnable temperature parameter. 

Accordingly, image-text contrastive loss $\mathcal{L}_{itc}$ is defined as: 
\begin{equation}
    \begin{split}
    \mathcal{L}_{itc}=\frac{1}{2}(\mathrm{CrossEntropy}(p^{i2t}, y^{i2t}) \\
    + \mathrm{CrossEntropy}(p^{t2i}, y^{t2i})),
    \end{split}
\label{eq:ITC}
\end{equation}
where $y^{i2t}, y^{t2i}\in \{0, 1\}^{M_Q}$ represents the label relation between the image and text. 
The positive image-text pair have value of $1$, while negative pairs being $0$.

After employing the explicit global alignment between $F^I$ and $F^T$ by Eq.~\ref{eq:ITC}, we feed $F^I$ and $F^T$ into the cross-modal encoder to obtain the image-text fusion representations $F^O=\{f^ O, f^O_{1}, f^O_{2}, \cdots , f^O_{L_T}\}\in \mathbb{R}^{(L_T+1)\times d}$, based on which we utilize a  linear projection matrix $W^O\in \mathbb{R}^{d\times 1}$ to calculate the fine similarity between the image and text:
\begin{equation}
    sim_{fine}(I, T)= f^O W^O,
    \label{eq:fine}
\end{equation}
where $\mathrm{sim}_{fine}(I, T)$ measures the similarity between global image representations and text after fusion. 

Based upon the fine similarity, the image-text matching loss can be formulated as:
\begin{equation}
    \mathcal{L}_{itm} = CrossEntropy(sim_{fine}(I, T), y^{itm}),
\end{equation}
where $y^{itm}$ denotes whether this image-text pair is a positive pair, $1$ for positive and $0$ for negative. Typically, the negative sample is sampled from the training batch. 

Although the image-text matching loss is effective in facilitating the training of the cross-modal encoder, it tends to prioritize a single pair and fails to account for the contrast between pairs. 
To be specific, the positive pair $(I, T)$ should output higher $\mathrm{sim}_{fine}(I, T)$ compared to the negative pairs $(I, T_{neg})$ and $(I_{neg}, T)$.
To satisfy it, In addition to $\mathcal{L}_{itc}$ and $\mathcal{L}_{itm}$ in the baseline, we add a fusion triplet loss, 
\begin{equation}
    \begin{aligned}
        \mathcal{L}_{tri} = \lbrack\mathrm{sim}_{fine}(I, T) - \mathrm{sim}_{fine}(I_{neg}, T) + \delta\rbrack_+^2 + \\
        \lbrack\mathrm{sim}_{fine}(I, T) - \mathrm{sim}_{fine}(I, T_{neg}) + \delta\rbrack_+^2,
    \end{aligned}
    \label{eq:tri}
\end{equation}
where $\delta$ is the least margin between positive pairs and negative ones, and $\lbrack\cdot\rbrack_+$ represents $\text{max}(\cdot, 0)$. 

\subsection{Candidate selection strategy}
In the inference phase, considering that one-by-one cross encoding's time complexity is $O(mn)$, where $m, n$ are the numbers of texts and images, respectively. For accelerating inference, we employ the candidate selection strategy, as ALBEF~\cite{li2021align} does. 
Given a text query is $T$, we compute the coarse similarity $sim_{coarse}(I, T)$ in Eq.~\ref{eq:coarse} with each gallery image, based on which we select top-$k$ images according to the descending order of the similarities and then input them to the cross-modal encoder.
Finally, the fine similarity $sim_{fine}(I, T)$ between the query text and the top-k images in Eq.~\ref{eq:fine} is adopted for text-based person search.

\subsection{Experiments}

\subsubsection{Datasets and Evaluation Metrics} \label{app:dataset}
\paragraph{Datasets}
We conduct experiments on three text-based person search datasets: CUHK-PEDES~\cite{li2017person}, ICFG-PEDES~\cite{ding2021semantically}, and RSTPReid~\cite{zhu2021dssl}.
Below we introduce each dataset's details.
\begin{itemize}
\item \textbf{CUHK-PEDES} is the most commonly used dataset for text-based person retrieval, which contains $40,206$ images of $13,003$ persons and $80,412$ textual descriptions. The official split divides the data into 34,054 images of 11,003 identities for training and 3,074 images of 1,000 identities for testing.

\item \textbf{ICFG-PEDES} contains $54,522$ images
and corresponding textual descriptions of 4,102 pedestrians, which are split into a training
set with $34,674$ images of $3,102$ pedestrians, as well as a testing
set with $19,848$ images of $1,000$ pedestrians.

\item \textbf{RSTPReid} is composed of $20,505$ images and $41,010$
textual descriptions with $4,101$ identities, among which
$3,701$, $200$ and $200$ identities are utilized for training, validation
and testing.
\end{itemize}

\paragraph{Evaluation Metrics} 
We adopt the Rank-k metric ($k=1,5,10$) to evaluate
the performance, which is referred to simply as R@k in the following sections. R@k metric represents the probability of at least one matching person image being in the top-$k$ candidate list given a query description. We also utilize mean Average Precision(mAP).
\begin{figure}[!htbp]
    \centering
    \setlength{\intextsep}{-1pt}
    \setlength{\abovecaptionskip}{-0.01cm}
    \setlength{\belowcaptionskip}{1cm}
    \includegraphics[width=\linewidth]{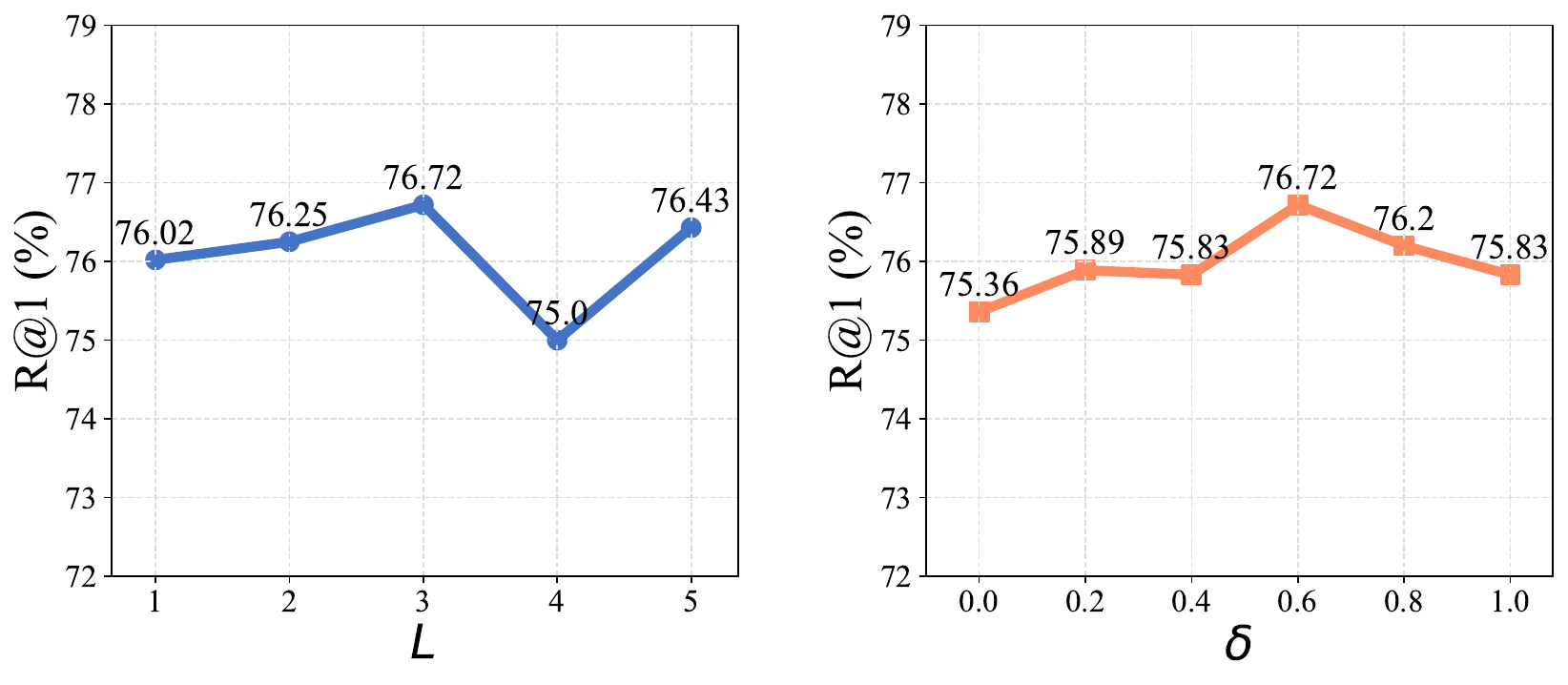}
    \caption{Analysis of the parameters. $L$ is the index of the cross-attention layer for generating BidirAtt weights, and $\delta$ is the margin in the fusion triplet loss.}
    \label{fig:parameters}
\end{figure}
\begin{figure*}[htbp]
    \centering
    \setlength{\intextsep}{-1pt}
    \setlength{\abovecaptionskip}{-0.01cm}
    \setlength{\belowcaptionskip}{1cm}
    \includegraphics[width=0.95\linewidth ]{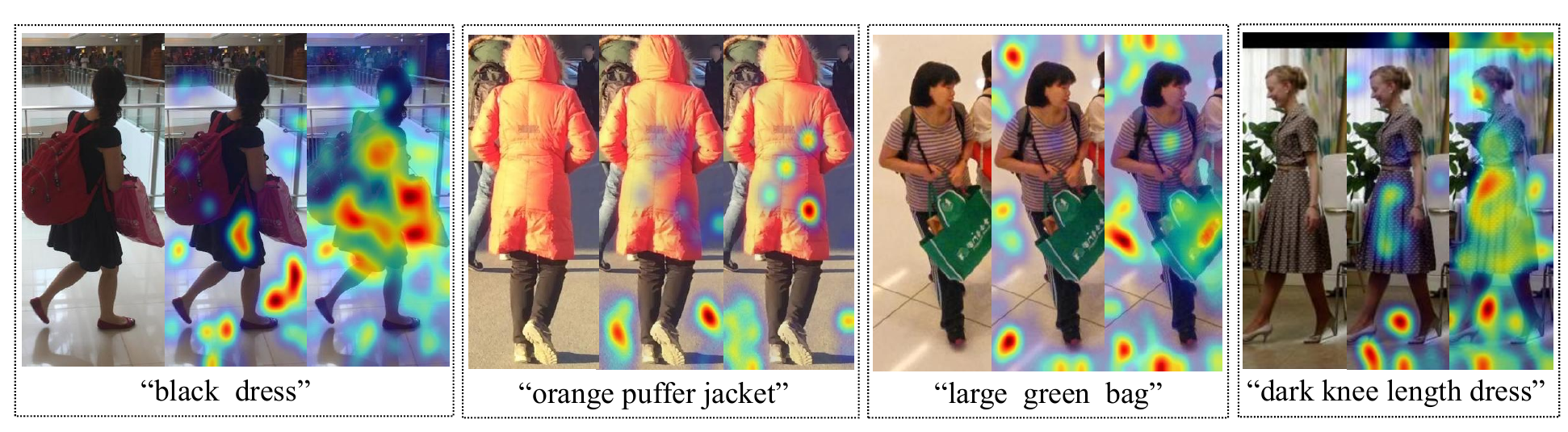}
    \caption{
     Visualization of the activation map of forward and bidirectional attention.
     For each group, we show \emph{the input image}, \emph{the activation map of ForAtt}, and \emph{the activation map of BidirAtt} from left to right.
     The brighter an area, the stronger its activation to the input phrase below.}
    \label{fig:visualization2}
\end{figure*}
\subsubsection{Implementation Details}\label{app:Implementation}
All experiments are conducted on 4 NVIDIA 3090 GPUs. For all images, we resize them to $384\times 384$ before inputting them into the model. For all texts, we limit the maximum input length to $50$ tokens, which means if a text has more than $50$ tokens, we only take its first $50$ tokens as input and discard other tokens, and if a text has fewer than $50$ tokens, we append $[PAD]$ tokens to its tail till its length reaches $50$. For both texts and images, the global and local representations have $768$ dimensions, and the projected representations for coarse similarity computation have $256$ dimensions.
We load the pretrained weights of ALBEF~\cite{li2021align} and further pre-finetune ALBEF on COCO~\cite{coco} dataset with $\mathcal{L}_{itc}$ and $\mathcal{L}_{itm}$ for $10$ epochs.
The proposed LAIP is trained on the dataset of text-based person search in a two-stage way.
In the first stage, it is trained for 30 epochs with only $\mathcal{L}_{itc}$ and $\mathcal{L}_{itm}$ losses. 
In the second stage, it is trained for 15 epochs with all losses. 
In both stages, we employ AdamW~\cite{loshchilov2018decoupled} as our optimizer and use cosine learning rate decay with an initial learning rate of $1\times10^{-5}$ and a warmup for the learning rate of $1\times10^{-6}$. The momentum coefficient $\alpha$ for updating the momentum model is configured to $0.995$. The margin between positive and negative pairs $\delta$ in $\mathcal{L}_{tri}$ is set to $0.6$. 

\subsubsection{Ablation Study}\label{app:analysis}

\textbf{Analysis of Which Cross-Attention Layer to Generate BidirAtt Weights.}
The forward and backward attentions in the BidirAtt can be computed via the output of any cross-attention layer.
We investigate the influence of different layers' output on performance. 
Specifically, there are $6$ layers of cross-attention in LAIP. The last layer has backward attention weights equalling 0, because the flow of gradients does not reach the sampled regions in the last cross-attention layer. Thereby we skip the $6$-th layer in our ablation study.
The results are shown in the left-hand side of Fig. \ref{fig:parameters}. 
LAIP performs best when the $3$-th layer is selected for generating forward and backward attentions. The reason can be that the $3$-th layer captures abundant fine-grained visual details, while higher layers tend to capture more semantic information and lower more details. 

\textbf{Analysis of the parameter $\delta$ of fusion triplet loss.}
The margin $\delta$ is a hyperparameter that controls the distance of positive and negative pair samples in $\mathcal{L}_{tri}$.
The influence of $\delta$ is shown in the right of Fig. \ref{fig:parameters}.
The results indicate that LAIP performs best when $\delta$ is set to 0.6. A smaller margin decreases the distance between positive and negative samples and could cause over-focusing on distinguishing between them, reducing the model's generalization ability and thus affecting performance. A larger margin increases the distance between positive samples, making the model overly conservative and challenging to capture subtle differences.
\begin{table}[!htbp]
    \centering
    \setlength{\intextsep}{-1pt}
    \setlength{\abovecaptionskip}{-0.01cm}
    \setlength{\belowcaptionskip}{-1cm}
    \caption{Effectiveness of the COCO Finetuning Strategy on CUHK-PEDES.}
    \resizebox{0.48\textwidth}{!}{
        \begin{tabular}{clc|cccc}
            \hline
            No. & Method & COCO-finetune & R@1 & R@5 & R@10 & mAP \\
            \hline
            1 & Baseline & \ding{55}  & 63.87 & 81.73 & 86.42 & 54.41 \\
            2 & Baseline$^+$ & \ding{51} & 66.59 & 84.31 & 88.99 & 57.08 \\
            \hline
            3 & LAIP & \ding{55}  & 72.30 & 86.71  & 90.46 & 62.07 \\
            4 & LAIP$^+$ & \ding{51} & \textbf{76.72} & \textbf{90.42}  & \textbf{93.60} & \textbf{66.05} \\
            \hline
        \end{tabular}
        }
    \label{tab:analysis_coco_finetune}
\end{table}

\textbf{Analysis of the COCO Pre-finetuning.}
The COCO pre-finetuning strategy helps to establish a buffer between pre-training task and the downstream text-based person search task. As shown in Table ~\ref{tab:analysis_coco_finetune}, a comparison between No.1 and No.2 reveals that pre-finetuning the baseline model on the COCO dataset yields a notable performance enhancement of 2.72\% in R@1 and 2.67\% in mAP. 
Furthermore, we analyze the impact of COCO pre-finetuning on the proposed LAIP. The results of No.3 and No.4 demonstrate that finetuning leads to a substantial improvement of 4.42\% in R@1 and 3.98\% in mAP.
These results effectively showcase the effectiveness of COCO pre-finetuning in enhancing the performance of the models.

\subsubsection{Visualization}

\textbf{Visualization of the activation maps of forward and bidirectional attentions.}
Fig.~\ref{fig:visualization2} illustrates qualitative comparisons between the forward and bidirectional attentions. The bidirectional attention provides more accurate activations for the object described by the input phrase. 
For the phrase ``dark knee length dress", the forward attention highlights the region near the hands and feet, which can reflect people's action (\emph{e.g.,} walking or carrying). This is useful for global alignment. The bidirectional attention also have emphasis on these region and obtains stronger activations where are highly correlated to the input phrase. The visualization demonstrates the effectiveness of the bidirectional attention.
\bibliographystyle{IEEEtranBST2/IEEEtranS}
\bibliography{IEEEtranBST2/IEEEfull}

\end{document}